\documentclass{article}
\usepackage{spconf,amsmath,graphicx}
\usepackage[ruled]{algorithm2e}
\makeatletter
\newcommand{\nosemic}{\renewcommand{\@endalgocfline}{\relax}}
\newcommand{\dosemic}{\renewcommand{\@endalgocfline}{\algocf@endline}}
\makeatother

\SetArgSty{textnormal}
\SetAlCapNameFnt{\small}
\SetAlCapFnt{\small}
\SetAlgorithmName{Algo}{Algo}{List of Algorithms}
\usepackage{algorithmic}
\usepackage{multicol}
\usepackage{multirow}
\usepackage{booktabs}
\usepackage{url}
\usepackage{todonotes}
\usepackage{caption}
\usepackage{subcaption}

\title{Stem-driven Language Models \\ for Morphologically Rich Languages}
\name{Yash Shah$^*$, Ishan Tarunesh$^*$, Harsh Deshpande, Preethi Jyothi
\thanks{$^*$Joint first authors}}
\address{Indian Institute of Technology Bombay, Mumbai, India}
\begin{document}
%
\maketitle
\begin{abstract}
Neural language models (LMs) have shown to benefit significantly from enhancing word vectors with subword-level information, especially for morphologically rich languages. This has been mainly tackled by providing subword-level information as an input; using subword units in the output layer has been far less explored. In this work, we propose LMs that are cognizant of the underlying stems in each word. We derive stems for words using a simple unsupervised technique for stem identification. We experiment with different architectures involving multi-task learning and mixture models over words and stems. 
We focus on four morphologically complex languages -- Hindi, Tamil, Kannada and Finnish -- and observe significant perplexity gains with using our stem-driven LMs when compared with other competitive baseline models. 
\end{abstract}
\section{ Introduction}
\label{sec:intro}

Language modeling is a fundamental problem in speech and language processing that involves predicting the next word given its context. Recurrent neural network language models (RNNLMs) have become the de facto standard for language modeling. They typically produce a next-word probability distribution over a fixed vocabulary of words. Such an approach has two main limitations. Word embeddings for infrequently occurring words in training data are poorly estimated. Also, predictions at the word level are largely immune to the subword structure in words. Both these limitations are exacerbated for morphologically rich languages in which words have numerous morphological variants, leading to large vocabularies where a significant fraction of words appear in the long tail of the word distribution. Leveraging subword information becomes especially important for such languages.  

In prior work, RNNLMs have typically exploited subword-level information at the input side and learn improved word embeddings by utilizing morpheme- and character-level information. \cite{vania}~present an exhaustive comparison of many such methods. Incorporating subword information within the output layer of RNNLMs has received less attention. We explore this direction and make the following specific contributions:

\vspace{-0.5em}
\begin{itemize}
\itemsep-0.3em
\item We present a new stem-based neural LM that predicts a mixture of stem probabilities and a mixture of word probabilities and meaningfully combines them. We also outline an unsupervised algorithm to identify stems of words.
\item We also present stem-based models that use multi-task learning and  consistently outperform their word-based counterparts.
\item We demonstrate the effectiveness of our proposed architecture by showing significant reductions in perplexities on four morphologically rich languages, Hindi, Tamil, Kannada and Finnish.
\item We provide a detailed analysis of the benefits of our stem-driven approach and also contrast our model with a control task that highlights the importance of stems.
\end{itemize}

\section{Model Description}
\begin{figure*}[t]
        \begin{subfigure}[b]{0.30\textwidth}
                \includegraphics[width=\linewidth]{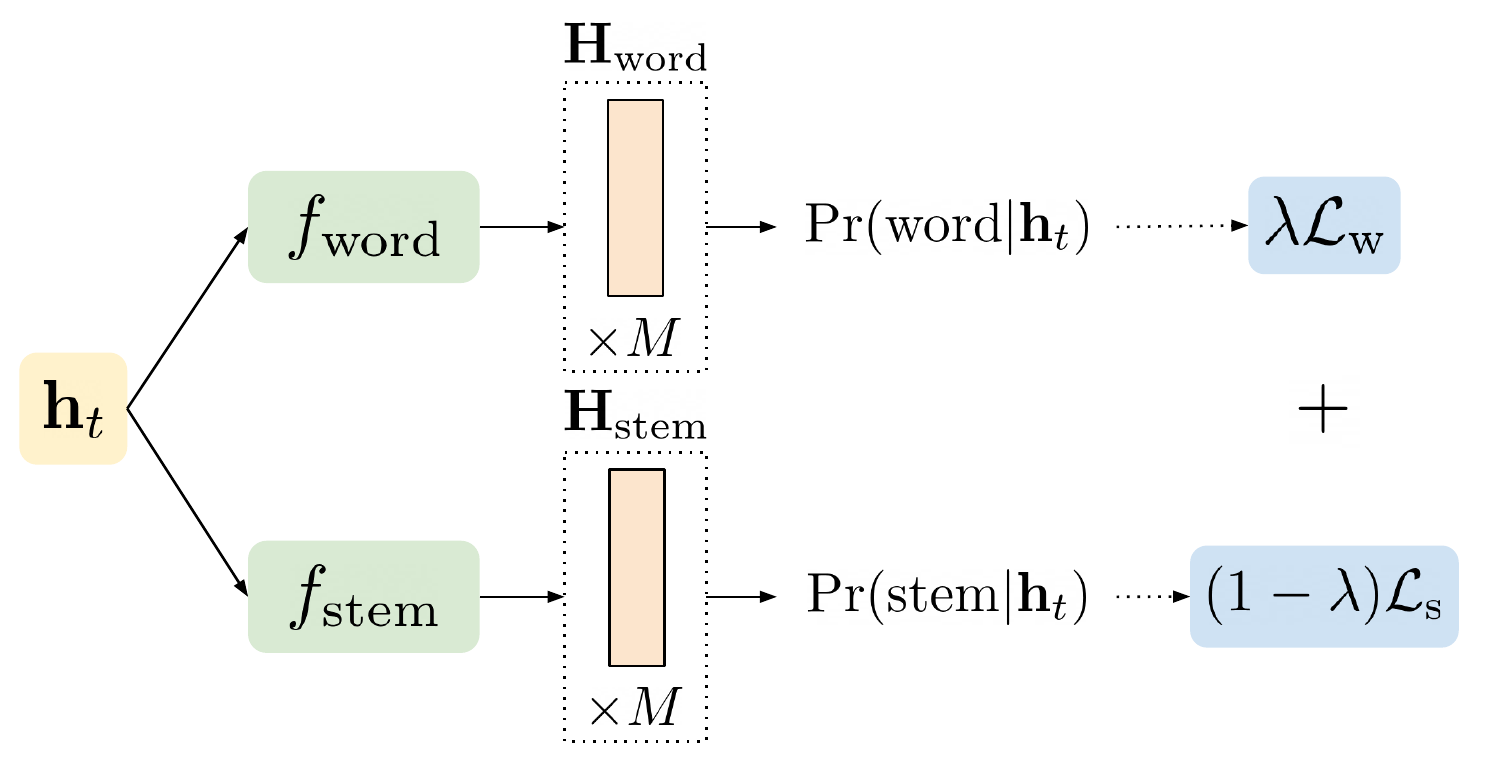}
                \caption{\textsc{MTL-S}}
                \label{fig:MTL-S}
        \end{subfigure}%
        \begin{subfigure}[b]{0.36\textwidth}
                \includegraphics[width=\linewidth]{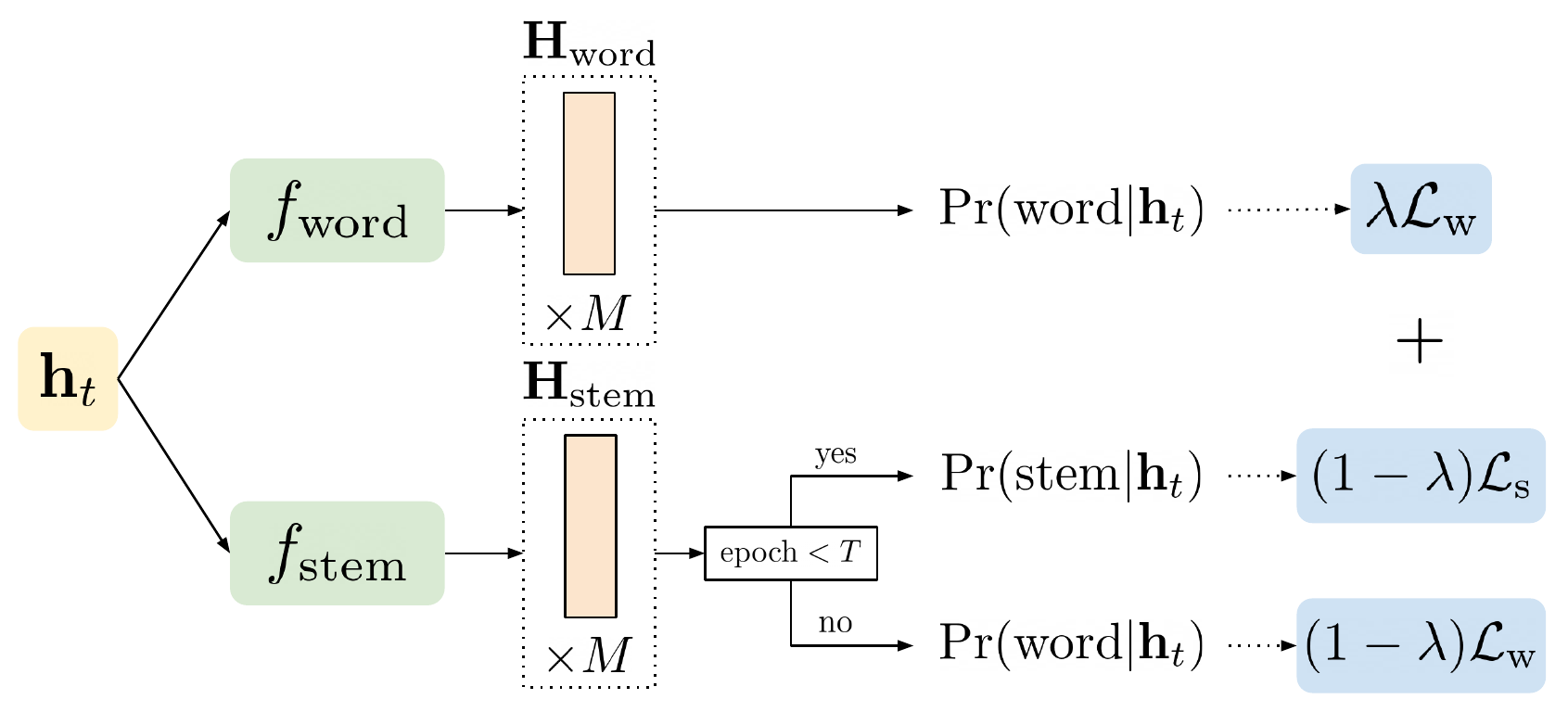}
                \caption{\textsc{MTL-S2W}}
                \label{fig:MTL-S2W}
        \end{subfigure}%
        \begin{subfigure}[b]{0.33\textwidth}
                \includegraphics[width=\linewidth]{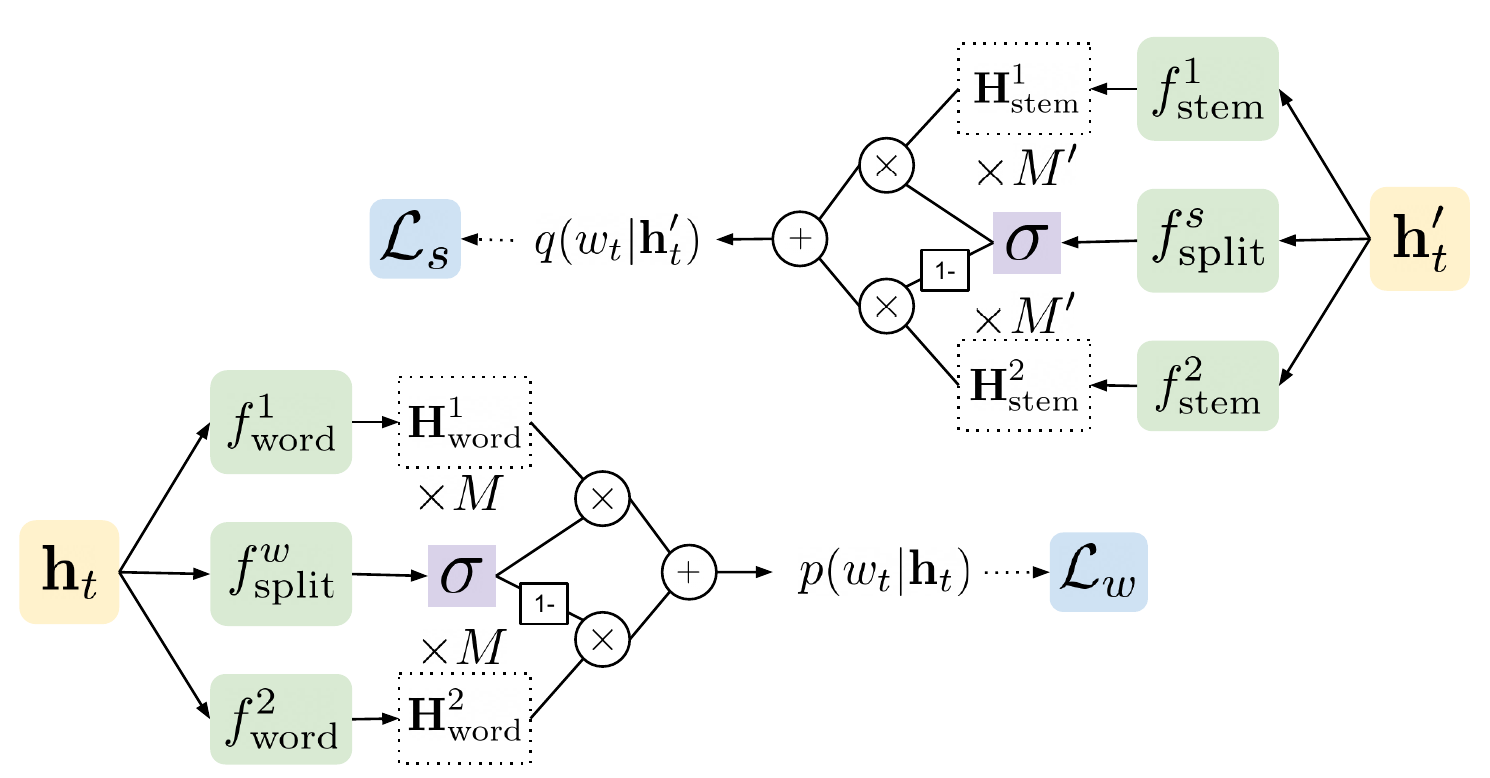}
                \caption{\textsc{Mix-WS}}
                \label{fig:tiger}
        \end{subfigure}
        \caption{Illustrating the workflow of our proposed models.}\label{fig:models}
\end{figure*}
\subsection{Preliminaries}

For a given word $w_t$, an RNNLM encodes its context $\{w_i\}_{i=1}^{t-1}$ into a fixed-size representation, $\mathbf{h}_t$. An RNNLM predicts $w_t$ by applying a softmax function to an affine transformation of $\mathbf{h}_t$ ($\mathbf{W}$ and $\mathbf{b}$ are the model parameters):
\setlength{\abovedisplayskip}{5pt}
\setlength{\belowdisplayskip}{5pt}
\setlength{\abovedisplayshortskip}{0pt}
\setlength{\belowdisplayshortskip}{0pt}
\begin{align}
P(w_t|\{w_i\}_{i=1}^{t-1}) = P(w_t|\mathbf{h}) = \mathrm{softmax}(\mathbf{W}\mathbf{h}_t+\mathbf{b})
\label{eqn:basiclm}
\end{align}

In a departure from this standard formulation, we could also use a mixture of $K$ different softmax distributions at the output layer. Compared to using a single softmax distribution, mixture model-based LMs lead to improved generalization abilities and translate into substantial reductions in test perplexities~\cite{neubig2016,matthews,yang}. If ${\lambda_t \in [1,K]}$ denotes the model at time $t$, the next-word distribution in a mixture model becomes:
\setlength{\abovedisplayskip}{0pt}
\setlength{\belowdisplayskip}{3pt}
\setlength{\abovedisplayshortskip}{0pt}
\setlength{\belowdisplayshortskip}{0pt}
\begin{align}
&P(w_t|\{w_i\}_{i=1}^{t-1}) = \sum_{k=1}^K P(w_t,\lambda_t=k|\mathbf{h}_t) \nonumber \\[-1em]
&= \sum_{k=1}^K P(\lambda_t=k|\mathbf{h}_t) P(w_t|\lambda_t=k,\mathbf{h}_t)
\label{eqn:mixturemodel}
\end{align}

RNNLMs are trained to minimize a cross-entropy loss function computed over all training tokens (indexed by $t=1,\ldots,T$), $\mathcal{L}_{w} = -\frac{1}{T}\sum_{t=1}^T\log P(w_t|\mathbf{h}_t)$ where $P(w_t|\mathbf{h}_t)$ can be estimated using a single softmax layer or a mixture of softmax layers as defined in Eqn~\ref{eqn:basiclm} or Eqn~\ref{eqn:mixturemodel}, respectively.

In all subsequent sections, we assume access to a stem for each word (which could be the word itself). We obtain this stem information using an unsupervised stemming algorithm that is detailed in Section~\ref{sec:segmentation}.

\subsection{Stem-based LMs using multi-task learning}
\label{sec:stem-mtl}

We treat the standard RNNLM defined in Eqn~\ref{eqn:basiclm} as the primary task of predicting words and augment it with an auxiliary task of predicting stems. Unlike the loss function $\mathcal{L}_w$ for the primary task that is computed over words, the auxiliary task is trained using a cross-entropy loss between the predicted stem and the correct stem $\mathcal{L}_s$. As is standard in multi-task learning, a linear combination of both these losses $\lambda \mathcal{L}_w + (1 - \lambda)\mathcal{L}_s$ will be optimized during training and $\lambda$ is a hyperparameter that we tune on a validation set. We refer to this model as \textbf{\textsc{MTL-S}}.
During test time, we discard the auxiliary task and only use the word probabilities from the primary task.

In \textsc{MTL-S}, the auxiliary task incurs no loss for predicting the correct stems but incurs a loss even if it predicts the correct word. In order to relax this constraint, we optimize the auxiliary task using $\mathcal{L}_s$ only for the first few epochs and use a word-level cross-entropy loss $\mathcal{L}_w$ for the remaining epochs. We refer to this model as \textbf{\textsc{MTL-S2W}}.

\subsection{\textsc{Mix-WS}: Using mixtures of words and stems}

We define a new language model, \textbf{\textsc{Mix-WS}}, that uses estimates from two mixture models -- one computed over words and another over stems -- to compute the final probability for a word given its context. We first train a mixture-model over words as defined in Eqn~\ref{eqn:mixturemodel}; let us call this distribution $p(w_t|\mathbf{h}_t)$. We also train a separate mixture-model over stems, $q(w_t|\mathbf{h}'_t)$. In this model, $w_t$ will correspond only to valid stems while $\mathbf{h}'_t$ encodes a context of past words; we estimate $q$ using a cross-entropy loss between the predicted stem and the correct stem.%
\footnote{We use the same vocabulary to represent both words and stems.}

Below we list the sequence of steps used to estimate word probabilities from \textsc{Mix-WS} for a test word $w$ appearing in a word context (encoded as $\mathbf{h}$ or $\mathbf{h}'$): 
\setlength{\abovedisplayskip}{5pt}
\setlength{\belowdisplayskip}{5pt}
\setlength{\abovedisplayshortskip}{0pt}
\setlength{\belowdisplayshortskip}{0pt}
\begin{enumerate}
\itemsep-0.4em
\item Compute the conditional probability of predicting $w$ given its stem $r(w) = \frac{p(w|\mathbf{h})}{p'_{\mathrm{stem}(w)}}$ where $p'_s = \displaystyle \sum_{w \in \mathcal{S}(s)} p(w)$ and $\mathcal{S}(s) = \{w \in \mathcal{V} \mid \mathrm{stem}(w)=s\}$.
\item Compute the marginal probability of predicting a stem $q'_s = \displaystyle \sum_{w \in \mathcal{S}(s)} q(w|\mathbf{h}')$.
\item The final probability for the test word $w$ given its context is computed as $r(w) \cdot q'_{\mathrm{stem}(w)}$.
\end{enumerate}

Figure~\ref{fig:models} illustrates our proposed models. Both $\textsc{MTL-S}$ and $\textsc{MTL-S2W}$ use fixed values of $\lambda$ to scale the loss terms specific to each task, while $\textsc{Mix-WS}$ uses learned weights (that change with the word context) to mix probabilities before computing the word- and stem-specific losses.
In the next section, we outline an unsupervised stem identification algorithm to derive stem-like entities from a word vocabulary.

\subsection{Unsupervised stem identification algorithm}
\label{sec:segmentation}


Algorithm~\ref{algo:segmentor} describes the pseudocode of our unsupervised segmentation method. For each prefix pair $(p_1, p_2)$ (and similarly for each suffix pair), we enforce $p_1<p_2$ which is true if ${|p_1|<|p_2|}$, or ${|p_1|=|p_2|}$ and $p_1$ is lexicographicaly smaller than $p_2$.
Frequently occurring pairs, guided by the threshold parameters $\delta_s$ and $\delta_p$, are chosen to form the set of prefix/suffix rules governing segmentation. Finally, for each word, the most frequent stem is returned as the output.

Several unsupervised word segmentation algorithms have been previously developed to discover the morphology of a language~\cite{goldsmith,creutz_morph,pitler,cotterell}. However, in this work, we did not examine the impact of changing the segmentation algorithm on LM performance and we leave this exploration for future work.

\begin{algorithm}[t!]
    \small
        \SetAlgoLined
        \KwIn{vocabulary $\mathcal{V}$, threshold parameters $\delta_s$ and $\delta_p$}
        \KwOut{function $\mathrm{stem}:\mathcal{V} \rightarrow \mathcal{V}$}

        \nosemic Let $\mathcal{R}_\text{p}$ be set of prefix pairs $(p_1,p_2)$ where $p_1=p_2=\epsilon$, \;
        \hfill or $p_1<p_2$ and there are at least $\delta_p$ pairs $(w_1,w_2) \in \mathcal{V}^2$ \;
        \dosemic \hfill s.t.\ $w_1=p_1+u$, $w_2=p_2+u$ for some $u$ \;

        \nosemic Let $\mathcal{R}_\text{s}$ be set of suffix pairs $(s_1,s_2)$ where $s_1=s_2=\epsilon$, \;
        \hfill or $s_1<s_2$ and there are at least $\delta_s$ pairs $(w_1,w_2) \in \mathcal{V}^2$ \;
              \dosemic \hfill s.t.\ $w_1=u+s_1$, $w_2=u+s_2$ for some $u$ \;

        \nosemic $\mathcal{R} \leftarrow  \{ (v, w)\in \mathcal{V}^2 | \exists (p_1,p_2) \in \mathcal{R}_\mathrm{p},
        (s_1,s_2) \in \mathcal{R}_\mathrm{s}$ \;
        \dosemic \hfill  and $u, $ s.t.\ $v = p_1 + u + s_1, w=p_2+u+s_2 \} $  \;

        \For{$v\in\mathcal{V}$}{
                $\mathrm{wt}[v]\leftarrow | \{ w \in \mathcal{V} \mid (v,w) \in \mathcal{R} \} |$ \;
        }

        \For{$w\in\mathcal{V}$}{
                $\displaystyle  \mathrm{stem}(w) := \operatorname*{arg\,max}_{v:(v,w) \in \mathcal{R} }(\mathrm{wt}[v])$ \;
        }
        \caption{Unsupervised stem identification}
        \label{algo:segmentor}
\end{algorithm}

\section{Experimental Setup}
\label{sec:expts}

\begin{table}[b!]
\centering
\setlength{\belowcaptionskip}{-10pt}
\begin{tabular}{l*4c}
\toprule
\multicolumn{1}{c}{\multirow{2}{*}{Statistic}} &  \multicolumn{4}{c}{Language} \\
 & Hi & Kn & Ta & Fi\\
\midrule
\# of training tokens & 666K & 434K & 507K &  585K\\
Vocabulary size & 50K & 94K & 106K & 115K\\
Type / Token (train) & 0.08 & 0.22 & 0.21 & 0.197\\
\# of dev tokens & 50K & 24K & 39K & 43K\\ 
\# of test tokens & 49K & 29K & 39K & 44K\\
OoV rate (test) & 5.3\% & 4.9\% & 15.2\%  & 5.6\% \\
\bottomrule
\end{tabular}
\caption{Dataset statistics.}
    \label{tab::stats}
\end{table}


\subsection{Dataset description}
In this work, we use datasets from~\cite{gerz} for four morphologically rich languages, Finnish (Fi), Hindi (Hi), Kannada (Kn) and Tamil (Ta), along with their specified training/dev/test splits. Table~\ref{tab::stats} shows statistics for these four languages. 
Kn, Ta and Fi are more morphologically complex than Hi, which is apparent from their higher type-token ratios in Table~\ref{tab::stats}. 



\subsection{Implementation details}
\label{sec:impl}

PyTorch~\cite{pytorch} was used to implement all models. We report two baseline numbers: (A) \textsc{Char-CNN-LSTM}: An RNNLM proposed by~\cite{kim} that uses character-level inputs for a variety of languages and (B) \textsc{LMMRL}: An RNNLM proposed by~\cite{gerz} that improves over~\cite{kim} by finetuning the output embeddings to capture subword-level information. We report numbers for (A) and (B) using our  re-implementations of these baseline systems, which are better than the reported numbers in~\cite{gerz} (except for Hi).%
\footnote{We also investigated BPE-based neural LMs as baselines. But these produced significantly worse test perplexities than \textsc{LMMRL} for all languages.}
Since our datasets were all relatively small in size, we present test perplexities averaged over five random seeds for all models.

We used SGD for the baseline models and ran each model for 30 epochs based on~\cite{gerz}. For all our proposed models, we used the Adam optimizer with the learning rate set to 5e-5 (decayed by 0.8) and ran each model for 15 epochs. The best value of $\lambda$ for our \textsc{MTL} models was found by tuning on the development set for a single random seed. \textsc{MTL-S2W} was trained for 5 epochs to optimize $\mathcal{L}_s$ and the remaining 10 epochs were used to optimize $\mathcal{L}_w$.
All hyperparameters like batch size, sequence length, embedding size, LSTM parameters were kept constant across both languages and models.

\begin{table*}[t!]
\centering
\setlength{\belowcaptionskip}{-10pt}
\begin{tabular*}{\hsize}{@{\extracolsep{\fill}} cccccc@{}}
\toprule
& & Hi & Kn & Ta & Fi\\
\midrule
\multicolumn{2}{c}{\textsc{Char-CNN-LSTM}}&
383.03 \begin{small}$\pm$6.96\end{small} &
1403.19 \begin{small}$\pm$58.05\end{small} &
2321.00 \begin{small}$\pm$180.64\end{small} &
1998.00 \begin{small}$\pm$90.25\end{small}\\
\multicolumn{2}{c}{\textsc{LMMRL}}&
375.51 \begin{small}$\pm$10.88\end{small} &
1404.22 \begin{small}$\pm$129.77\end{small} &
2241.00 \begin{small}$\pm$240.80\end{small} &
2017.63 \begin{small}$\pm$88.21\end{small}\\
\midrule
\multicolumn{2}{c}{\textsc{MTL-W}}&
305.43 \begin{small}$\pm$17.51\end{small} &
971.40 \begin{small}$\pm$65.45\end{small} &
1567.03 \begin{small}$\pm$193.36\end{small} &
1322.81 \begin{small}$\pm$46.48\end{small}\\
\multicolumn{2}{c}{\textsc{MTL-S}}&
311.08 \begin{small}$\pm$21.18\end{small} &
946.58 \begin{small}$\pm$57.69\end{small} &
1529.65 \begin{small}$\pm$168.07\end{small} &
1411.16 \begin{small}$\pm$88.94\end{small}\\
\multicolumn{2}{c}{\textsc{MTL-S2W}}&
304.24 \begin{small}$\pm$20.91\end{small} & 
918.45 \begin{small}$\pm$48.39\end{small} &
1489.92 \begin{small}$\pm$155.60\end{small} &
1374.81 \begin{small}$\pm$68.56\end{small}\\
\midrule
\multicolumn{2}{c}{\textsc{Mix-W}}&
284.91 \begin{small}$\pm$18.76\end{small} & 
795.93 \begin{small}$\pm$100.74\end{small} &
1178.22 \begin{small}$\pm$322.61\end{small} &
1112.46 \begin{small}$\pm$282.13\end{small}\\
\multicolumn{2}{c}{\textsc{Mix-WS}}&
265.27 \begin{small}$\pm$23.68\end{small} & 
764.76 \begin{small}$\pm$119.03\end{small} &
1160.43 \begin{small}$\pm$224.41\end{small} &
1133.48 \begin{small}$\pm$130.89\end{small}\\
\bottomrule
\end{tabular*}
\caption{Average perplexities (with standard deviations) on Hi, Kn, Ta and Fi test sets.}
    \label{tab::results}
\end{table*}

\section{Results and Analysis}

\subsection{Test perplexities of the proposed models}

Table~\ref{tab::results} summarizes test perplexities averaged over five random seeds on Hi, Kn, Ta and Fi. We present two new models, \textsc{MTL-W} and \textsc{Mix-W} to serve as fair comparisons to \textsc{MTL-S}, \textsc{MTL-WS} and \textsc{Mix-WS}. In \textsc{MTL-W}, both the primary and auxiliary tasks predict words and only the primary task is used at test time. \textsc{Mix-W} estimates a mixture model computed over words as defined in Eqn~\ref{eqn:mixturemodel}.     

\begin{figure}[t!]
    \centering
    \includegraphics[width=0.8\linewidth]{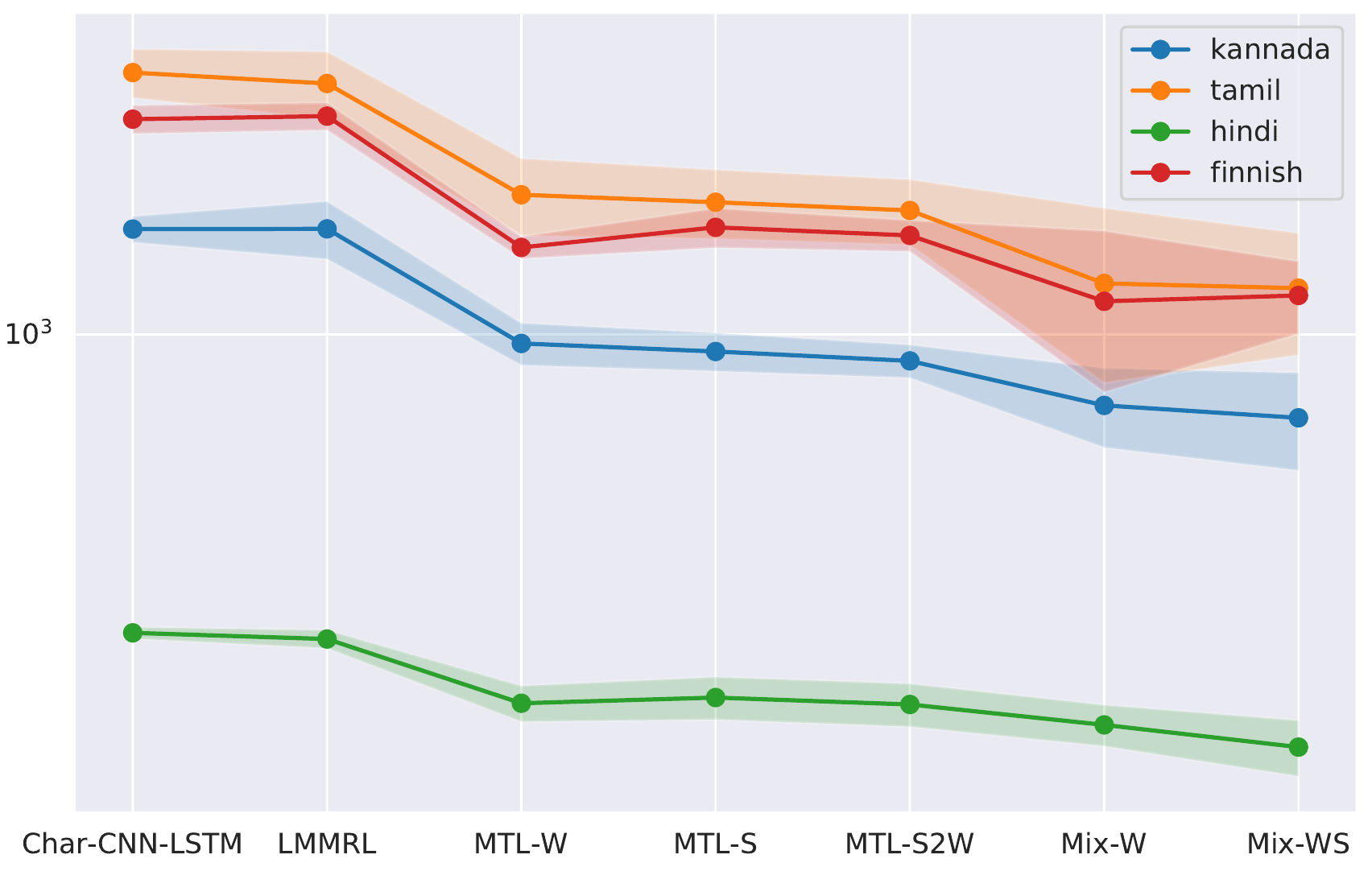}
    \caption{Perplexity trends across languages for all models.}
    \label{fig:ppl_trend}
\end{figure}

We observe that all our proposed models significantly outperform the two baselines, \textsc{Char-CNN-LSTM} and \textsc{LMMRL}, on all four languages. (On Kn and Ta, test perplexities are essentially halved.) \textsc{MTL-W} is almost always worse than \textsc{MTL-S} and \textsc{MTL-S2W} produces consistently lower perplexities than \textsc{MTL-S}. This suggests that incorporating stem information in the auxiliary task as a pretraining step (in \textsc{MTL-S2W} where the auxiliary task is trained first with a stem-based loss followed by a word-based loss) is beneficial to the primary word prediction task. The mixture models, \textsc{Mix-WS} and \textsc{Mix-W}, have significantly lower perplexities compared to all other models. \textsc{Mix-WS} also  consistently outperforms \textsc{Mix-W} (except for Fi%
\footnote{Fig~\ref{fig:ppl_trend} shows the perplexity trends across models for all four languages. Although \textsc{Mix-W} performs slightly better than \textsc{Mix-WS} on average for Fi, the variance of \textsc{Mix-W} is significantly larger than for any other model.}%
), further validating the importance of using a mixture-model for stems during test time. (Since \textsc{Mix-WS} has roughly twice the number of parameters in \textsc{Mix-W}, we trained a \textsc{Mix-W} model for Kn that was comparable in size. This performed worse than the \textsc{Mix-W} model listed in Table~\ref{tab::results} giving a perplexity of 830.938\begin{small}$\pm 83.06$\end{small}.)
%

\subsection{Control task to assess importance of stems}
\label{sec:control}

We set up a control task in order to assess the utility of stems identified by our unsupervised algorithm. First, we started with the list of stems  for a word vocabulary from our algorithm. Next, we randomly assigned words to be associated with each of these stems. We were careful to assign the same number of words to a stem as in our algorithm so as to not alter its distribution. With this randomized word-to-stem assignment in place for Kn, we run our best model \textsc{Mix-WS}. We obtain test perplexities of 1161.91 $\begin{small}\pm 194.53\end{small}$ and 764.76 $\begin{small}\pm 119.03 \end{small}$  using the randomized stem assignment and the stem assignments from our algorithm, respectively. This clearly shows that our derived stems are useful abstractions of the underlying words.

\subsection{Supervised vs. Unsupervised stems}

We hypothesise that our proposed models will perform even better if the quality of stems are further improved. In order to empirically validate this claim, we use a supervised segmentor for Finnish~\cite{omorfi}
%
to produce segments for each word which were then merged using a frequency-based criterion\footnote{The split was chosen by maximising the sum of resulting stem and suffix frequencies, inversely weighted by their global averages. We fixed the number of suffixes to 75 (same as that for our algorithm) to ensure a fair comparison.} to generate the stem and suffix. With these stems in place, we obtain an averaged test perplexity of 1095.60 $\begin{small}\pm 208\end{small}$ using the \textsc{Mix-WS} model, compared to 1133.48 $\begin{small}\pm 130.89\end{small}$ using our unsupervised algorithm to generate stems. 

\begin{figure}
    \centering
    \includegraphics[width=0.8\linewidth]{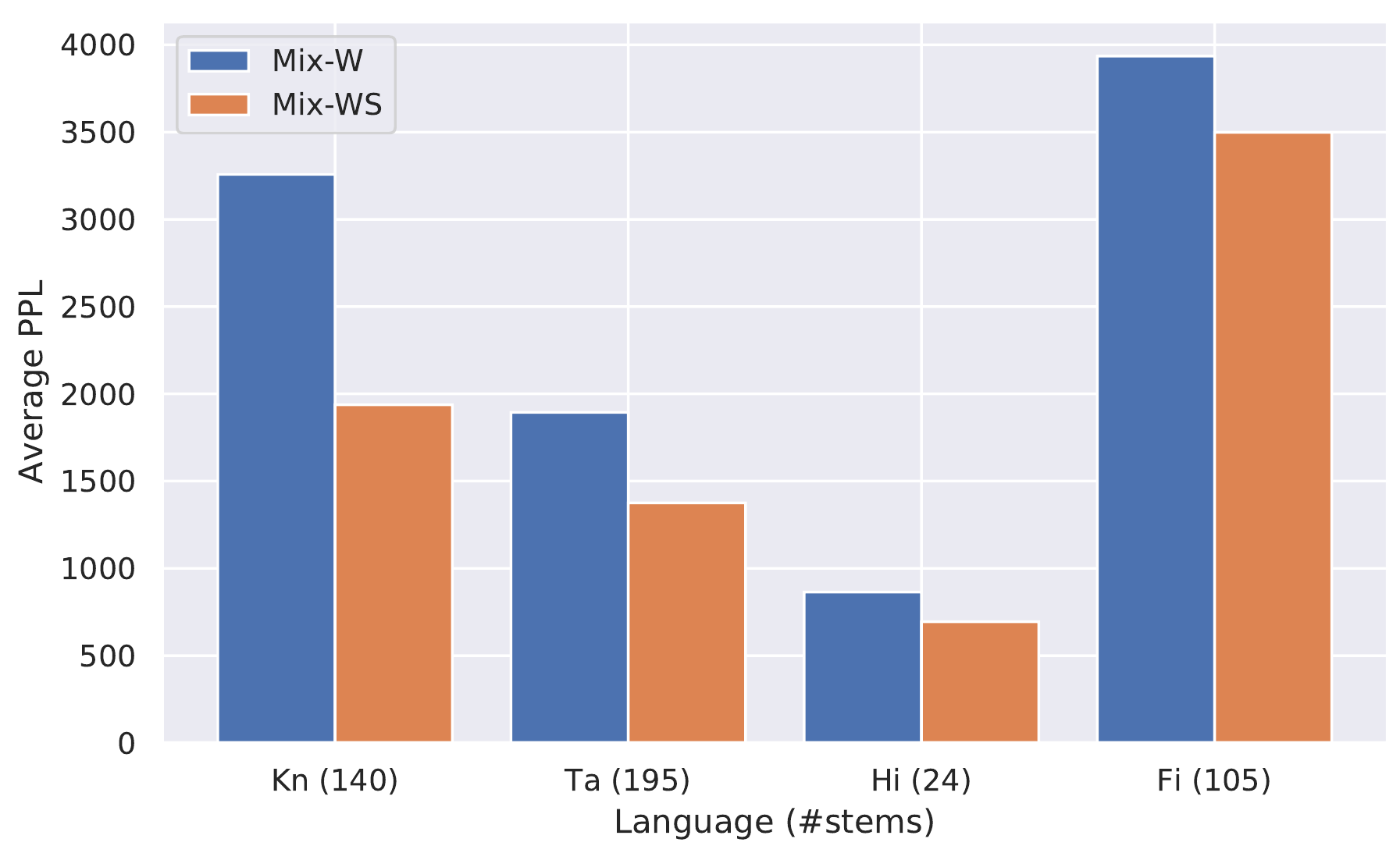}
    \caption{\textsc{Mix-W} and ${\textsc{Mix-WS}}$ on more frequent stems.}
    \label{fig:bar_plot}
\end{figure}

\subsection{Token-level perplexities for frequent and diverse stems} 

We compared the performance of the \textsc{Mix-W} and \textsc{Mix-WS} models on stems with sufficient coverage which have diverse word forms. We isolated stems that had 10 or more distinct word types that mapped to it, and these word tokens collectively appeared 500 or more times in the training data. We computed averaged test perplexities for only these tokens; Fig~\ref{fig:bar_plot} shows these values. We see consistent improvements on these specific tokens; the gap between \textsc{Mix-WS} and \textsc{Mix-W} is much larger for these tokens for Kn and Ta. 




\section{Related Work}

Prior work has looked at different ways in which morpheme or character-level information can be provided as input to RNNLMs~\cite{santos,kim,ling,lample}. Approaches tailored specifically for morphologically rich languages include using constituent morpheme embeddings~\cite{botha_blunsom}, using morphological recursive NNs~\cite{luong}, concatenating word and character embeddings~\cite{verwimp} and using other factored representations of words~\cite{vylomova,ataman,labeau,cao-rei}. Factored RNNLMs that integrate multiple word features (POS tags, etc.) have also been explored in prior work~\cite{wu,hoang}. Fewer approaches have focused on injecting subword-level information into the output layer of neural LMs. \cite{gerz}~proposed a finetuning technique for word embeddings using a loss based on character-level similarities. 
\cite{yuret}~and~\cite{varjokallio} split words into subwords and trained an LM using subwords as tokens and \cite{matthews} used a mixture model to predict at the word, morpheme and character-level.


\section{Conclusions}

In this work, we present stem-driven LMs for different morphologically rich languages and demonstrate their efficacy compared to competitive baseline models. We derive stems using a simple unsupervised technique and demonstrate how our models' performance can be further improved with predicting better stems. In future work, we will examine the effect of different segmentation algorithms on LM performance.

\newpage

\bibliography{ICASSP}
\bibliographystyle{IEEEbib}
\end{document}